\pdfoutput=1

\documentclass[11pt]{article}

\usepackage{naacl2021}

\usepackage{times}
\usepackage{latexsym}

\usepackage[T1]{fontenc}

\usepackage[utf8]{inputenc}

\usepackage{microtype}

\usepackage{subcaption}
\usepackage{graphicx}
\usepackage{enumitem}
\usepackage{microtype}
\usepackage{adjustbox}
\usepackage{algorithm}
\usepackage{algpseudocode}
\usepackage{multirow}

\title{Extractive and Abstractive Explanations \\for Fact-Checking and Evaluation of News}

\author{
    Ashkan Kazemi, Zehua Li, Ver\'{o}nica Per\'{e}z-Rosas and Rada Mihalcea\\
    \{ashkank, simonli, vrncapr, mihalcea\}@umich.edu\\
    University of Michigan, Ann Arbor \\
    }

\begin{document}

\maketitle
\begin{abstract}
In this paper, we explore the construction of natural language explanations for news claims, with the goal of assisting fact-checking and news evaluation applications.
We experiment with two methods: (1) an extractive method based on Biased TextRank -- a resource-effective unsupervised graph-based algorithm for content extraction; and (2) an abstractive method based on the GPT-2 language model. 
We perform comparative evaluations on two misinformation datasets in the political and health news domains, and find that the extractive method shows the most promise.
\end{abstract}

\section{Introduction}
Navigating the media landscape is becoming increasingly challenging given the abundance of misinformation, which reinforces the importance of keeping our news consumption focused and informed.
While fake news and misinformation have been a recent focus of research  studies~\cite{perez-rosas-etal-2018-automatic, thorne-vlachos-2018-automated, lu-li-2020-gcan}, the majority of this work aims to categorize claims, rather than generate explanations that support or deny them. This is a challenging problem that has been mainly tackled by expert journalists who manually verify the information surrounding a given claim and provide a detailed verdict based on supporting or refuting evidence. More recently, there has been a growing interest in creating computational tools able to assist during this process by providing supporting explanations for a given claim based on the news content and context~\cite{atanasova-etal-2020-generating, fan-etal-2020-generating}. While a true or false veracity label does not provide enough information and a detailed fact-checking report or news article might take long to read, bite-sized explanations can bridge this gap and improve the transparency of automated news evaluation systems.

To contribute to this line of work, our paper explores two approaches to generate supporting explanations to assist with the evaluation of news. First, we investigate how an extractive method based on Biased TextRank~\cite{kazemi-etal-2020-biased} can be used to generate explanations. Second, we explore an abstractive method based on GPT-2, a large generative language model \cite{radford2019language}. 

Our methods take as input a news article and a claim and generate a claim-focused explanation by extracting or generating relevant information to the original article in relation to the claim.  
We evaluate our proposed methods on the health care and political domains, where misinformation is abundant. As current news on the COVID-19 pandemic and the elections are overloading social media outlets, we find these domains to be of timely importance. 
Through comparative experiments, we find that both methods are effective at generating explanations for news claims, with the extractive approach showing the most promise for this task. 

\begin{table}[h]
\small
\begin{center}
\begin{tabular}{| p{0.95\columnwidth} |} 
\hline
\textbf{Claim:} nearly half of Oregon's children are poor. \\
\hline
\textbf{Fact-Check Report: }...Jim Francesconi...said..."Nearly half of Oregon's children are poor." He said the information came from a 2012 report...According to that report, "nearly 50\% of children are either poor or low-income." Francesconi almost immediately realized his mistake. "In retrospect, I wish I would have said poor or low income."...there is a distinction between poor and low income as far as the U.S. government is concerned." ... If you check the...Census information, you'll find that...23 percent of children in Oregon live in...below...poverty level while another 21 percent live in low-income families. As far as the U.S. government is concerned, about a quarter of the state's children are poor, not half... (redacted) \\
\hline
\textbf{Supporting Explanation (Ground Truth): }So where does this leave us? Francesconi said in an opinion piece that "nearly half of Oregon's children are poor." In fact, if you use federal definitions for poverty, about a quarter are poor and another quarter are low-income. But experts tell us that families that are described as low-income still struggle to meet their basic needs and, for all intents and purposes, qualify as poor. Be that as it may, Francesconi was referencing a report that used the federal definitions. \\
\hline
\textbf{Biased TextRank (Extractive): }"Nearly half of Oregon's children are poor." According to that report, "nearly 50\% of children are either poor or low-income." Low income refers to families between 100 and 200 percent of the federal poverty level. As far as the U.S. government is concerned, about a quarter of the state's children are poor, not half. \\
\hline
\textbf{GPT-2 Based (Abstractive):}" That's still below the federal poverty level. But that's not half. About 47 percent of Oregon's children are not poor, according to the Census data. So the percentage of children in the state who are poor is not half yet. It's actually closer to half. \\
\hline
\end{tabular}
\caption{An example data point from the LIAR-PLUS dataset, with  ground truth explanations, and explanations generated by our methods.}
\vspace*{-10pt}
\label{liar_example}
\end{center}
\end{table}

\section{Related Work}\label{sec:related work}
While explainability in AI has been a central subject of research in recent years~\cite{poursabzi2018manipulating, lundberg2017unified, core2006building}, the generation of natural language explanations is still relatively understudied. ~\citet{NIPS2018_8163} propose e-SNLI, a natural language (NL) inference dataset augmented with human-annotated NL explanations. In their paper,~\citeauthor{NIPS2018_8163} generated NL explanations for premise and hypothesis pairs for an inference task using the InferSent~\cite{conneau-etal-2017-supervised} architecture. ~\citet{kumar-talukdar-2020-nile} propose the task of generating ``faithful'' (i.e., aligned with the model's internal decision making) NL explanations and propose NILE, a method that jointly produces NLI labels and faithful NL explanations.

Generating explanations in the context of news and fact-checking is a timely and novel topic in the NLP community \cite{atanasova-etal-2020-generating, fan-etal-2020-generating, kotonya-toni-2020-explainable-automated}. 
In \cite{atanasova-etal-2020-generating} the authors proposed a supervised BERT~\cite{devlin-etal-2019-bert} based model for jointly predicting the veracity of a claim by extracting supporting explanations from fact-checked claims in the LIAR-PLUS~\cite{alhindi-etal-2018-evidence} dataset. \citet{kotonya-toni-2020-explainable-automated} constructed a dataset for a similar task in the public health domain and provided baseline models for explainable fact verification using this dataset. \citet{fan-etal-2020-generating} used explanations about a claim to assist fact-checkers and showed that explanations improved both the efficiency and the accuracy of the fact-checking process.

\section{Methods}\label{sec:methods}
We explore two methods for producing natural language explanations: an extractive unsupervised method based on Biased TextRank, and an abstractive method based on GPT-2. 
\subsection{Extractive: Biased TextRank}\label{sec:btextrank}
Introduced by~\citet{kazemi-etal-2020-biased} and based on the TextRank algorithm \cite{mihalcea2004textrank}, Biased TextRank is a targeted content extraction algorithm with a range of applications in keyword and sentence extraction. The TextRank algorithm ranks text segments for their importance by running a random walk algorithm on a graph built by including a node for each text segment (e.g., sentence), and drawing weighted edges by linking the text segment using a measure of similarity. 

The Biased TextRank algorithm takes an extra ``bias'' input and ranks the input text segments considering both their own importance and their relevance to the bias term. 
The bias query is embedded into Biased TextRank using a similar idea introduced by~\citet{haveliwala2002topic} for topic-sensitive PageRank. The similarity between the text segments that form the graph and the ``bias'' is used to set the restart probabilities of the random walker in a run of PageRank over the text graph. That means the more similar each text segment is to the bias query, the more likely it is for that node to be visited in each restart and therefore, it has a better chance of ranking higher than the less similar nodes to the bias query. During our experiments, we use SBERT~\cite{reimers-gurevych-2019-sentence} contextual embeddings to transform text into sentence vectors and cosine similarity as similarity measure.

\subsection{Abstractive: GPT-2 Based}\label{sec:gpt2}
We implement an abstractive explanation generation method based on GPT-2, a transformer-based language model introduced in~\citet{radford2019language} and trained on 8 million web pages containing 40 GBs of text. 

Aside from success in language generation tasks~\cite{budzianowski-vulic-2019-hello, ham-etal-2020-end}, the pretrained GPT-2 model enables us to generate abstractive explanations for a relatively small dataset through transfer learning.

In order to generate explanations that are closer in domain and style to the reference explanation, we conduct an initial fine-tuning step.
While fine tuning, we provide the news article, the claim, and its corresponding explanation as an input to the model and explicitly mark the beginning and the end of each input argument with bespoke tokens. At test time, we provide the article and query inputs in similar format but leave the explanation field to be completed by the model.
We use top-k sampling to generate explanations. We stop the generation after the model outputs the explicit end of the text token introduced in the fine-tuning process.

Overall, this fine-tuning strategy is able to generate explanations that follow a style similar to the reference explanation.
However, we identify cases where the model generates gibberish and/or repetitive text, which are problems previously reported in the literature while using GPT-2~\cite{holtzman2019curious,welleck2020consistency}. 
To address these issues, we devise a strategy to remove unimportant sentences that could introduce noise to the generation process. We first use Biased TextRank to rank the importance of the article sentences towards the question/claim. Then, we repeatedly remove the least important sentence (up to 5 times) and input the modified text into the GPT-2 generator.
This approach keeps the text generation time complexity in the same order of magnitude as before and reduces the generation noise rate to close to zero.
\section{Evaluation}\label{sec:generation}

\subsection{Experimental Setup}\label{evaluation:healthnewsreview:automatic}

We use a medium (355M hyper parameters) GPT-2 model~\cite{radford2019language} as implemented in the Huggingface transformers~\cite{Wolf2019HuggingFacesTS} library. 
We use ROUGE~\cite{lin-2004-rouge}, a common measure for language generation assessment as our main evaluation metric for the generated explanations and report the F score on three variations of ROUGE: ROUGE-1, ROUGE-2 and ROUGE-L.

We compare our methods against two baselines. The first is an explanation obtained by applying TextRank on the input text. The second, called ``embedding similarity'', ranks the input sentences by their embedding cosine similarity to the question and takes the top five sentences as an explanation.

\subsection{Datasets}\label{sec:datasets}
\paragraph{LIAR-PLUS.}\label{sec:datasets:liar}
The LIAR-PLUS~\cite{alhindi-etal-2018-evidence} dataset contains 10,146 train, 1,278 validation and 1,255 test data points collected from PolitiFact.com, a political fact-checking website in the U.S. A datapoint in this dataset contains a claim, its  verdict, a news-length fact-check report justifying the verdict and a short explanation called ``Our ruling'' that summarizes the fact-check report and the verdict on the claim. General statistics on this dataset are presented in Table~\ref{tab:statisticsDatasets}.

\paragraph{Health News Reviews (HNR).}\label{sec:datasets:hnr}

We collect health news reviews along with ratings and explanations from healthnewsreview.org, a website dedicated to evaluating healthcare journalism in the US.~\footnote{We followed the restrictions in the site's \textit{robots.txt} file.} The news articles are rated with a 1 to 5 star scale and the explanations, which justify the news' rating, consist of short answers for 10 evaluative questions on the quality of information reported in the article. The questions cover informative aspects that should be included in the news such as intervention costs, treatment benefits, discussion of harms and benefits, clinical evidence, and availability of treatment among others. Answers to these questions are further evaluated as either satisfactory, non-satisfactory or non-applicable to the given news item. For our experiments, we select 1,650 reviews that include both the original article and the accompanying metadata as well as explanations. Explanations' statistics are presented in Table~\ref{tab:statisticsDatasets}. 

To further study explanations in this dataset, we randomly select 50 articles along with their corresponding questions and explanations. We then manually label sentences in the original article that are relevant to the quality aspect being measured.\footnote{The annotation was conducted by two annotators, with a Pearson's correlation score of 0.62 and a Jaccard similarity of 0.75. 
} During this process we only include explanations that are deemed as ``satisfactory,'' which means that relevant information is included in the original article.

\begin{table}
\small
\begin{tabular}{lllll}
\hline
 Dataset & Total count & Av. Words & Av. Sent. \\ \hline
LIAR-PLUS & 12,679 & 98.89 & 5.20 \\
HNR & 16,500 & 87.82 & 4.63 \\ \hline
\end{tabular}%

\caption{Dataset statistics for explanations; total count, average words and sentences per explanation.}
\label{tab:statisticsDatasets}
\end{table}

\begin{table}[t]
\small
\begin{tabular}{ l  c  c  c } 
\hline
 
Model & ROUGE-1 & ROUGE-2 & ROUGE-L \\
\hline
TextRank & 27.74 & 7.42 & 23.24 \\
GPT-2 Based & 24.01 & 5.78 & 21.15 \\
Biased TextRank & \textbf{30.90} & \textbf{10.39} & \textbf{26.22} \\
\hline
\end{tabular}
\caption{ROUGE-N scores of generated explanations on the LIAR-PLUS dataset.
}
\label{table:liar_plus}
\end{table}

\begin{table*}[h]
\small
\begin{center}
\begin{tabular}{ l  c  c  c  c  c  c } 
\hline
\multirow{2}{4em}{Model} & \multicolumn{3}{c}{Explanations} & \multicolumn{3}{c}{Relevant Sentences} \\
 & ROUGE-1 & ROUGE-2 & ROUGE-L & ROUGE-1 & ROUGE-2 & ROUGE-L \\
\hline
Embedding Similarity & 18.32 & 2.96 & 15.25 & 22.02 & 8.79 & 20.21 \\ 
GPT-2 Based & \textbf{20.02} & \textbf{4.32} & \textbf{17.67} & 15.74 & 2.58 & 13.32 \\
Biased TextRank & 19.41 & 3.41 & 15.87 & \textbf{23.54} & \textbf{10.15} & \textbf{21.88} \\
\hline
\end{tabular}
\caption{ROUGE evaluation on the HNR dataset. Left columns under ``Explanations'' have the actual explanations as reference and the columns on the right provide results for comparison against question-relevant sentences.
}
\label{table:hnr:automatic}
\end{center}
\end{table*}

\begin{table}[h]
\small
\centering
\begin{tabular}{ l    c  c  c } 
\hline
Model & Acc. & F1 (+) & F1 (-) \\
\hline
GPT-2 Based & 64.40\% & 49.04\% & 54.67\% \\
Biased TextRank & \textbf{65.70\%} & \textbf{56.69\%} & \textbf{57.96\%} \\
\hline
\end{tabular}
\caption{Downstream evaluation results on the HNR dataset, averaged over 10 runs and 9 questions.} 
\label{table:qa}
\end{table}

\subsection{Producing Explanations}
We use the Biased TextRank and the GPT-2 based models to automatically generate explanations for each dataset. With LIAR-PLUS, we seek to generate the explanation provided in the ``Our ruling'' section. For HNR we aim to generate the explanation provided for the different evaluative questions described in section~\ref{sec:datasets:hnr}. We use the provided training, validation and test splits for the LIAR-PLUS dataset. For HNR, we use 20\% of the data as the test set and we study the first nine questions for each article only and exclude question \#10 as answering it requires information beyond the news article. We use explanations and question-related article sentences as our references in ROUGE evaluations over the HNR dataset, and the section labeled ``Our ruling'' as ground truth for LIAR-PLUS.

\paragraph{Extractive Explanations.} To generate extractive explanations for the LIAR dataset, we apply Biased TextRank on the original article and its corresponding claim and pick the top 5 ranked sentences as the explanation (based on the average length of explanations in the dataset).
To generate explanations on the HNR dataset, we apply Biased TextRank on each news article and question pair for 9 of the evaluative questions and select the top 5 ranked sentences as the extracted explanation (matching the dataset average explanation length).

\paragraph{Abstractive Explanations.}
We apply the GPT-2 based model to generate abstractive explanations for each dataset using the original article and the corresponding claim or question as an input. We apply this method directly on the LIAR-PLUS dataset. On the HNT dataset,  since we have several questions, we train separate GPT-2 based models per question. In addition, each model is trained using the articles corresponding to questions labeled as ``satisfactory'' only as the ``unsatisfactory" or ``not applicable" questions do not contain information within the scope of the original article.

\subsection{Downstream Evaluation}
We also conduct a set of experiments to evaluate to what extent we can answer the evaluation questions in the HNR dataset with the generated explanations. For each question, we assign binary labels to the articles (1 for satisfactory answers, 0 for not satisfactory and NA answers) and  train individual classifiers aiming to discriminate between these two labels. During these experiments each classifier is trained and evaluated ten times on the test set and the results are averaged over the ten runs. 

\section{Experimental Results}\label{sec:generation:results}
As results in Table~\ref{table:liar_plus} suggest, while our abstractive GPT-2 based model fails to surpass extractive baselines on the LIAR-PLUS dataset, Biased TextRank outperforms the unsupervised TextRank baseline. Biased TextRank's improvements over TextRank suggest that a claim-focused summary of the article is better at generating supporting explanations than a regular summary produced by TextRank. Note that the current state-of-the-art results for this dataset, presented in \cite{atanasova-etal-2020-generating} achieve 35.70, 13.51 and 31.58 in ROUGE-1, 2 and L scores respectively.
However, a direct comparison with their method would not be accurate as it is a method that is {\it supervised} (versus  the unsupervised Biased TextRank) and {\it extractive} (versus the  abstractive GPT-2 based model).

Table ~\ref{table:hnr:automatic} presents results on automatic evaluation of generated explanations for the HNR dataset, showing that the GPT-2 based model outperforms Biased TextRank when evaluated against actual explanations and Biased TextRank beats GPT-2 against the extractive baseline.
This indicates  the GPT-2 based method is more effective in this dataset and performs comparably with Biased TextRank. Results for the downstream task using both methods are shown in Table~\ref{table:qa}. As observed, results are significantly different and demonstrate that Biased TextRank significantly outperforms (t-test $p=0.05$) the GPT-2-based abstractive method, thus suggesting that Biased TextRank generates good quality explanations for the HNR dataset. 
\section{Discussion} 
Our evaluations indicate that Biased TextRank shows the most promise, while the GPT-2 based model mostly follows in performance. Keeping in mind that the GPT-2 based model is solving the harder problem of {\it generating} language, it is worth noting the little supervision it receives on both datasets, especially on the HNR dataset where the average size of the training data is 849. 

In terms of resource efficiency and speed, Biased TextRank is faster and lighter than the GPT-2 based model. Excluding the time needed to fine-tune the GPT-2 model, it takes approximately 60 seconds on a GPU to generate a coherent abstractive explanation on average on the LIAR-PLUS dataset, while Biased TextRank extracts explanations in the order of milliseconds and can even do it without a GPU in a few seconds. We find Biased TextRank's efficiency as another advantage of the unsupervised algorithm over the GPT-2 based model.

\section{Conclusion}\label{sec:conclusion}
In this paper, we presented extractive and abstractive methods for generating supporting explanations for more convenient and transparent human consumption of news. We evaluated our methods on two domains and found promising results for producing explanations. In particular, Biased Text-Rank (an extractive method) outperformed the unsupervised baselines on the LIAR-PLUS dataset and performed reasonably close to the  extractive ground-truth on the HNR dataset.

For future work, we believe generating abstractive explanations should be a priority, since intuitively an increase in the readability and coherence of the supporting explanations will result in improvements in the delivery and perception of news.

\section*{Acknowledgments}
We are grateful to Dr. Stacy Loeb, Professor of Urology and Population Health at New York University, for her expert feedback, which was instrumental for this work. 
This material is based in part upon work supported by the Precision Health initiative at the University of Michigan, by the National Science Foundation (grant \#1815291), and by the John Templeton Foundation (grant \#61156). Any opinions, findings, and conclusions or recommendations expressed in this material are those of the author and do not necessarily reflect the views of the Precision Health initiative, the National Science Foundation, or John Templeton Foundation.

\bibliography{References,anthology}
\bibliographystyle{acl_natbib}

\end{document}